# Automatic Induction of Bellman-Error Features for Probabilistic Planning
# Online Appendix 1

**Jia-Hong Wu**                                                             JW@ALUMNI.PURDUE.EDU
**Robert Givan**                                                            GIVAN@PURDUE.EDU
*Electrical and Computer Engineering,*
*Purdue University, W. Lafayette, IN 47907 USA*

## 1. Additional Pseudo-code and Grammar

In this section, we present additional pseudo-code and feature grammar for our feature-learning framework as follows:

1. Pseudo-code for our trajectory-based approximate value iteration (AVI) approach is shown in Figure 1 on page 2.

2. Pseudo-code for drawing training sets by following a policy is shown in Figure 2 on page 2.

3. Our relational feature grammar is shown in Figure 3 on page 3.

## 2. Details on the Selection and Modification of Competition Domains

Every goal-oriented domain with a problem generator from the first or second IPPC (Younes, Littman, Weissman, & Asmuth, 2005; Bonet & Givan, 2006) was considered for inclusion in our experiments. For inclusion, we require a planning domain with a fixed action definition, as defined in Section 2.4 in the paper, that in addition has only ground conjunctive goal regions. Four domains have these properties directly, and we have adapted three more of the domains to have these properties as we describe in the next paragraph. The resulting selection provides seven IPPC planning domains for our empirical study. Figure 4 lists the reasons for the exclusion of the other six goal-oriented domains. In addition, four of the domains that we use in evaluation occur in both competitions in slightly different forms and we evaluate on one version of each of these four, as described in Figure 5.

    The three domains we adapted for inclusion are as follows. We created our own problem generators for the first IPPC domains TOWERS OF HANOI and FILEWORLD, as none were provided in the competition. For both these domains, there is only one instance of each size. In Towers of Hanoi, all instances share the same action set and state predicates, so that a suitable problem generator is straightforward. In Fileworld, a planning domain with a fixed action definition results if we consider the collection of instances that share the same fixed number of folders, but varying the number of files. When the number of folders varies, the state predicates and actions change, so that instances with varying numbers of folders cannot be in the same fixed-action-definition planning do-

**AVI**

| | |
|---|---|
| Inputs: | Feature vector $\overrightarrow{\Phi} = (f_0, \ldots, f_n)$, weight vector $\overrightarrow{w} = (w_0, \ldots, w_n)$ |
| Outputs: | Weight vector $\overrightarrow{w}$ |

1.     $\overrightarrow{w}^1 \leftarrow \overrightarrow{w}$.
2.     **for** $\beta \leftarrow 1$ **to** $T_{\text{AVI}}$
3.         $(s_1, \ldots, s_m) \leftarrow \mathbf{draw}(\text{Greedy}(\overrightarrow{w}^\beta \cdot \overrightarrow{\Phi}), N_{\text{AVI}})$.
4.         **for** $i \leftarrow 0$ **to** $n$
5.             $n_i \leftarrow |\{j|f_i(s_j) \neq 0\}|$.
6.             $w_i^{\beta+1} = w_i^\beta + \frac{1}{n_i} \sum_{j=1}^{m} \alpha f_i(s_j)(\mathcal{U}(\overrightarrow{w}^\beta \cdot \overrightarrow{\Phi})(s_j) - (\overrightarrow{w}^\beta \cdot \overrightarrow{\Phi})(s_j))$.
7.     **return** $\overrightarrow{w}^{T_{\text{AVI}}+1}$.

Notes:

1. $T_{\text{AVI}}$ and $N_{\text{AVI}}$ are system parameters.
2. $\mathbf{draw}(\pi, n)$ draws a sequence of $n$ states following $\pi$.
3. $\mathcal{U}(V)$ is the Bellman update of $V$, as defined in Section 2.1 in the paper.
4. $\alpha$ is the learning rate.

Figure 1: Pseudo-code for approximate value iteration (AVI).

**draw**

| | |
|---|---|
| Inputs: | Policy $\pi$, number of states to be drawn $n$ |
| Outputs: | Sequence of states $\Delta$ |

1.     $s \leftarrow$ MDP initial state $s_0$, $\Delta \leftarrow ()$, $j \leftarrow 0$.
2.     **for** $i \leftarrow 1$ **to** $n$
3.         $\Delta \leftarrow (\Delta; s)$, $j \leftarrow j + 1$.
4.         $s \leftarrow$ sample next state from $S$ according to $T(s, \pi(s), \cdot)$.
5.         **if** $s = \bot$ **or** $j >$ maximum training trajectory length
6.             $s \leftarrow$ MDP initial state $s_0$, $j \leftarrow 0$.
7.     **return** $\Delta$.

Figure 2: Pseudo-code for drawing training sets by following a policy.

main under our definitions (preventing natural generalization between sizes). For our experiments, we create a suitable domain by coding a problem generator restricted to three folders.

Furthermore, FILEWORLD, as written for the competition, is partially propositionalized (for unknown reasons). First, rather than have a one-argument predicate "have-folder", the competition domain has one proposition "have-$f$" for each folder $f$. Also, the competition domain duplicates

| | | |
|---|---|---|
| ⟨term⟩ | ::= | ⟨variable⟩ \| ⟨constant⟩ |
| ⟨goal-based-enrichment⟩ | ::= | goal-⟨domain-predicate⟩ \| correct-⟨domain-predicate⟩ |
| ⟨predicate⟩ | ::= | ⟨domain-predicate⟩ \| ⟨goal-based-enrichment⟩ |
| ⟨enriched-predicate⟩ | ::= | ⟨predicate⟩ \| ⟨predicate⟩+ \| min-⟨predicate⟩ \| max-⟨predicate⟩ |
| ⟨atom⟩ | ::= | ⟨enriched-predicate⟩ (⟨term$_1$⟩, ⋯ , ⟨term$_n$⟩), where $n$ is the arity of ⟨enriched-predicate⟩ |
| ⟨literal⟩ | ::= | ⟨atom⟩ \| ¬ ⟨atom⟩ |
| ⟨conjunction⟩ | ::= | ⟨literal⟩ \| ⟨conjunction⟩ ∧ ⟨literal⟩ |
| ⟨feature-expression⟩ | ::= | ⟨conjunction⟩ \| ∃ ⟨variable⟩ ⟨feature-expression⟩ |

Notes:

1. A *free variable* in a feature expression is a variable that is not inside the scope of a quantifier (∃) for that variable.
2. A *feature* is a feature-expression with at most one free variable.
3. ⟨domain-predicate⟩ is given by the planning domain definition.
4. ⟨goal-based-enrichment⟩ is taken to be null for non-goal-oriented planning domains.
5. +, min-, and max-, in the production for ⟨enriched-predicate⟩, can only be applied to ⟨predicate⟩ expressions of arity 2.
6. Arity is extended to enriched predicates as follows:
   (a) + produces an enriched predicate of arity 2.
   (b) min- and max- produce enriched predicates of arity 1.
   (c) goal-⟨predicate⟩ and correct-⟨predicate⟩ have the same arity as ⟨predicate⟩.

Figure 3: A grammar for our relational feature language.

and renames each action for each folder rather than take a folder object as an action argument (again for unknown reasons). Finally, the competition domain contains an apparent bug because it does not give types to the objects, so it is possible to file a folder in itself. Because we study relational generalization here, we have constructed the obvious lifted version of this domain with object types; we include the PPDDL source as Section 4 of this appendix. We call the resulting domain LIFTED-FILEWORLD3.

Finally, for BOXWORLD, we modify the problem generator so that the goal region is always a ground conjunctive expression by replacing the goal "all boxes must be at their destinations" with a conjunction of specific box location goals. We call the resulting domain CONJUNCTIVE-BOXWORLD.

## 3. Parameterization of Our Methods

Here we describe our choice of parameters for our methods. Where possible, parameterization is done once, to apply identically to all experiments, as described here. There are some choices made once for each domain, and these are described in the subsection dedicated to each domain. The primary choices that must be made in a domain-specific way control learning from small problems: we must specify for each domain the performance threshold at which difficulty will be increased (as shown in Figure 1 in the paper) as well as the sequence of difficulties to be considered (in cases where there is more than one parameter controlling problem size). We defer to future research the topic of automated control of problem difficulty when learning from small problems. We currently make these choices by experimentation with the domain; our experience with such experimentation suggests that these choices can successfully be automated in the future.

### 3.1 Trajectory Termination

Training sets for both feature learning and for AVI weight update are drawn by drawing trajectories based on the current greedy policy in problems drawn from the problem distribution at the current level of difficulty, as detailed in Sections 3 and 2.5 in the paper. It is an important and somewhat independent research topic to automatically recognize when such a trajectory is not making progress, e.g., by recognizing dead-end regions of states and/or lack of progress towards the goal. Any such research can be plugged into our methods directly by terminating all training trajectories when they fail an appropriate test.

Here, we do not address this issue in any sophisticated way, but terminate trajectories whenever one of three conditions holds:

1. a goal state is reached,

2. a dead-end state is reached,

3. the trajectory contains 1,000 steps.

| Domain name | IPPC version | Reason for exclusion |
|---|---|---|
| Colored blocksworld | IPPC1 | Goal region is not a ground conjunction |
| Drive | IPPC2 | Uses predicates with three or more arguments |
| Elevators | IPPC2 | Uses predicates with three or more arguments |
| Pitchcatch | IPPC2 | Action definition not fixed throughout domain |
| Schedule | IPPC2 | Action definition not fixed throughout domain |
| Random | IPPC2 | Action definition not fixed throughout domain |

Figure 4: Reasons for excluding some planning competition domains from our experiments.

| Domain name | Differences | Version used | Reason for choice |
|---|---|---|---|
| Blocksworld | Many small differences<br>– IPPC2 adds **emptyhand**, **on-table**$(x)$, and **clear**$(x)$<br>– IPPC2 removes table object<br>– IPPC2 adds actions: **pick-up-from-table**, **put-down**, **pick-tower**, **put-tower-on-block**, and **put-tower-down**<br>– IPPC2 allows **on**$(x,x)$ | IPPC1 | IPPC2 version inaccuracy allows **on**$(x,x)$ |
| Exploding blocks | No generator in IPPC1 | IPPC2 | Problem generator in IPPC2 |
| Tireworld | No generator in IPPC1 | IPPC2 | Problem generator in IPPC2 |
| Zenotravel | No generator in IPPC1 | IPPC2 | Problem generator in IPPC2 |

Figure 5: Differences between IPPC1 and IPPC2 versions of planning domains present in both competitions, which version is used in our experimental evaluation, and why.

### 3.2 Training Set Sizes

Each feature-learning training set across all our relational-learning experiments is drawn to be 20,000 states by the method described in Section 3 in the paper. Because propositional feature learning is faster than relational feature learning, we are able to allow 200,000 states in propositional feature learning training sets in the TETRIS and SYSADMIN experiments, but still only 20,000 states in the planning domains.

Throughout all experiments, each AVI weight-update training set is drawn by collecting the states from 30 trajectories.

### 3.3 Learning Rate for Weight Updates in AVI

As discussed in Section 2.5 in the paper, we adjust the weights of our approximated value functions using AVI. We use a search-then-converge schedule for the learning rate of this iterative gradient descent method throughout our experiments (see Darken & Moody, 1992); specifically, we set the learning rate $\alpha$ in AVI to $\frac{3}{1+k/100}$, where $k$ is the number of AVI iterations already executed.

### 3.4 Parametrization of the Relational Algorithm

There are various parameters in the feature construction process described in this section, including the beam-width $W$, the beam-search depth limit $d$, the regularization parameter $\lambda$, and the bound on the maximum number of quantifiers in scope $q$. Changes to these parameters affect the quality of the constructed features by changing the feature-space regions searched and the number of candidate features considered, as well as changing the preferences expressed in scoring the features. The selection of these parameters further affects the choice of the size of feature training set, as in practice fewer training examples can be considered when the number of candidate features grows.

Throughout all our experiments we choose $W$ to be 60, $d$ to be 5, and $\lambda$ to be 0.03 for all domains. We set $q$ to 1 for the planning competition domains (setting $q$ to 2 does not result in a noted improvement in the performance in these domains when using the above parameters, but results in a substantial and occasionally intolerable runtime cost), and we set $q$ to 2 for TETRIS. These severe limits on $q$ are necessary to control the expense of searching the feature space. Note however that there is implicit quantification in the transitive-closure predicates and min/max predicates in the extended predicate set defining the feature space, in addition to the explicit quantifiers limited by $q$. See Section 4.1 in the paper for discussion of the extended predicate set.

### 3.5 Parametrization of the Propositional Algorithm

Our propositional feature learning algorithm is already well defined in Section 4.4 in the paper, except for how to setup the underlying C4.5 learner (Quinlan, 1993). We use the default parameters for C4.5, except for the following: we use the gain criterion instead of the gain ratio criterion. We allow the trees to grow from a node without any restriction on the minimum number of objects in the resulting branches[1]. The pruning confidence level is set to 0.9.

## 4. PPDDL Source for Lifted-Fileworld3

The PPDDL source for LIFTED-FILEWORLD3 with a problem size of 10 files.

```
(define (domain file-world)
        (:requirements  :typing
                        :disjunctive-preconditions
                        :negative-preconditions
                        :conditional-effects
                        :probabilistic-effects
                        :universal-preconditions)
        (:types  file folder)

        (:predicates  (has-type ?p - file)
                      (goes-in ?p - file ?f - folder)
                      (filed ?p - file)
                      (have ?f - folder))
        (:constants  F0 F1 F2 - folder )

(:action get-type
        :parameters  (?p - file)
        :precondition  (and (not (has-type ?p)))
        :effect  (and (has-type ?p)
                      (probabilistic
                         0.333   (goes-in ?p F0)
```

---

1. The default C4.5 parameter requires at least 2 branches from any node to contain at least 2 objects.

```
                              0.333    (goes-in ?p F1)
                              0.334    (goes-in ?p F2))))

(:action get-folder
        :parameters  (?f - folder)
        :precondition  (and (forall (?x -folder) (not (have ?x))))
        :effect   (have ?f))

(:action file-F
        :parameters  (?p - file ?f - folder)
        :precondition  (and (have ?f) (has-type ?p)
                            (goes-in ?p ?f))
        :effect   (filed ?p))

(:action return-folder
        :parameters  (?f - folder)
        :precondition  (have ?f)
        :effect   (not (have ?f)))
)

(define (problem file-prob)
        (:domain file-world)
        (:objects p0 p1 p2 p3 p4 p5 p6 p7 p8 p9 )
        (:goal (and (filed p0) (filed p1) (filed p2) (filed p3)
                    (filed p4) (filed p5) (filed p6) (filed p7)
                    (filed p8) (filed p9)))
)
```

## 5. Modifications to the Weight Update Rule in AVI

**Scaling step-size during AVI** For the complex domains addressed in this paper, simple gradient descent has many potential pitfalls. One such pitfall is that the Bellman error surface may be extremely steep at some points. Because the weight changes in AVI are proportional to the gradient, arbitrarily large gradients result in arbitrarily large single-step weight changes that are rarely desirable (and can also cause floating-point overflow). There is a substantial literature on dynamically adjusting step size during gradient descent (Jacobs, 1988; Kwong & Johnston, 1992; Harris, Chabries, & Bishop, 1986; Mathews & Xie, 1993); however, gradient descent is not the main topic of this paper and so we resort only to a simple work-around for arbitrarily large gradients: rather than step proportional to the gradient, we compress the unbounded space of possible step sizes to a finite interval using a sigmoidal function, as described next. Large gradients here are due to large statewise Bellman error averages over the training set, as can be seen by examining the weight update equation, Equation 1, in Section 2.5 in the paper. Here we compress large weight updates by a sigmoidal scaling of the average statewise Bellman error, as described formally in the next three equations:

$$B_{avg} = \frac{1}{n} \sum_j (\mathcal{U}(V^\beta)(s_j) - V^\beta(s_j))$$

$$\kappa = \frac{1}{1 + \exp(-4(1 - |B_{avg}|/r_{\text{scale}}))}$$

$$w_i^{\beta+1} = w_i^\beta + \kappa \frac{1}{n_i} \sum_j \alpha f_i(s_j)(\mathcal{U}(V^\beta)(s_j) - V^\beta(s_j))$$

In our experiments, we use this approach to computing $w^{\beta+1}$ rather than the direct approach given by Equation 1. The scaling factor $\kappa$ will be close to one unless the average statewise Bellman error $B_{avg}$ grows large, and thus significant differences between the direct approach and the scaled approach appear only in that case. The sigmoidal function is a somewhat arbitrary choice here; any bounded, smooth, monotone function that is linear for a scalable-sized region near the origin will suffice. The domain-specific parameter $r_{\text{scale}}$ represents the reward scaling of the problem domain. We note that any MDP problem can be rescaled by multiplying all rewards by the same positive scalar with consequent rescaling of the value of any policy at any state by the same scalar. Our method here is not invariant to this rescaling and thus requires a hand-set domain parameter to represent the reward scaling. We select $r_{\text{scale}}$ using trial-and-error in each domain by starting from $r_{\text{scale}} = 1$, which suffices for all domains we evaluate here except SysAdmin, where we us $r_{\text{scale}} = 10$. We leave for future research the topic of automatically, possibly dynamically, finding the value of the reward scaling parameter.

**Sign restriction in weight adjustment** Another pitfall in using gradient descent with complex gradient surfaces is that dramatic increases in error can result from one step of weight update. In our AVI setting, this can result in dramatic drops in the success rate of the resulting greedy policy. Because in goal-oriented domains a useful gradient is computed only from successful trajectories, such dramatic drops in success rate can result in an uninformative gradient from which AVI often cannot recover. Various mechanisms can be designed for detecting dramatic drops in policy quality during AVI and revisiting the weight updates that lead to them; here we focus only on revisiting weight updates that change the sign of a weight, and only when the immediately resulting policy performs much worse than the policy before the weight update.

It is fairly intuitive that weight updates changing the sign of a weight are particularly suspect. If the weight for a feature has been tuned to a positive value, it is hopefully because that feature has been seen to correlate to the desired value function; however, this immediately implies that the negation of that feature anti-correlates with the desired value. Changing the sign of a weight is a form of rejecting previous training regarding the entire direction of the importance of the corresponding feature. Empirically, we have found that AVI on complex error surfaces often makes damaging mistakes by stepping too far in weight update to the degree that the sign of a feature is reversed and the resulting policy is suddenly severely degraded.

In our experiments in goal-oriented planning problems, we implement a mechanism to detect and avoid weight sign changes that must be avoided to preserve policy quality, as follows. First, we define a method for empirically comparing policies: we say that a policy $\pi_1$ "tests as significantly better" than a policy $\pi_2$ if Student's t-test confirms the hypothesis that the success rate of $\pi_2$ is at most 0.9 times the success rate of $\pi_1$ with significance 0.025 based upon 100 sample trajectories of each. Second, each time we construct an AVI training set by drawing trajectories, we measure the success rate of the policy Greedy($V$) used over the trajectories drawn to create the training set—we call this the training success rate of the value function $V$. If the training success of the current value function $V_2$ is lower than the training success of the previous value function $V_1$, we then test if the the policy Greedy($V_1$) tests as significantly better than the policy Greedy($V_2$). If so, we reconsider any weight sign changes (including changes to or from zero) made during the intervening weight update as follows. Suppose that $V_1$ is described by weights $w^\beta$ and $V_2$ by weights $w^{\beta+1}$. For each weight $w_i$ that changed sign from $w_i^\beta$ to $w_i^{\beta+1}$, we test if reversing the update of just that weight, using $w_i^\beta$ in place of $w_i^{\beta+1}$, yields a greedy policy that tests significantly better than Greedy($V_2$). Any such weights that yield significant improvements when their $\beta+1$-iteration updates are reversed are then restored to their $\beta$-iteration values and their sign is locked for the remainder of this run of AVI. In other words, any future weight update to that weight which would change the sign of that weight is replaced with no change to that weight.